\PassOptionsToPackage{svgnames}{xcolor}
\documentclass[sigconf, nonacm]{acmart}
\usepackage{multirow}
\usepackage{lscape}
\usepackage{scalefnt}
\usepackage{graphicx}
\usepackage{hhline}
\usepackage{pgfplots}
\usepackage{pgfplotstable}
\usepackage{algorithm2e}
\usepackage{ circuitikz }
\pgfplotsset{compat=1.18} 

\AtBeginDocument{%
  }

\begin{document}

\title{Jointly Generating and Attributing Answers using Logits of
Document-Identifier Tokens}
\thanks{Preprint, under review}
\author{Lucas Albarede$^1$, Jose G. Moreno$^1$, Lynda Tamine$^1$, Luce Lefeuvre$^2$}
\email{{lucas.albarede, jose.moreno, lynda.lechani}@irit.fr}
\email{{luce.lefeuvre}@sncf.fr}
\affiliation{%
  \institution{$^1$Université de Toulouse, IRIT}
  \city{Toulouse}
  \country{France}
}
\affiliation{%
  \institution{$^2$Dir. Technologies Innovation, SNCF}
  \city{Paris}
  \country{France}
}


\begin{abstract}
Despite their impressive performance, Large Language Models (LLMs) remain prone to hallucination, which critically undermines their trustworthiness. 
While most of the previous work focused on tackling answer and attribution correctness, a recent line of work investigated faithfulness, with a focus on leveraging internal model signals to reflect the model’s actual decision-making process while generating the answer. Nevertheless, these methods induce additional latency and have shown limitations in directly aligning token generation with attribution generation. In this paper, we introduce \textbf{LoDIT}, a method that jointly generates and faithfully attributes answers in RAG by leveraging specific token logits during generation. It consists of two steps: (1) marking the documents  with specific token identifiers and then leveraging the logits of these tokens to estimate the contribution of each document to the answer during generation, and (2) aggregating these contributions into document attributions. 
Experiments on a trustworthiness-focused attributed text-generation benchmark, Trust-Align, show that \textbf{LoDIT} significantly outperforms state-of-the-art models on several metrics. Finally, an in-depth analysis of \textbf{LoDIT} shows both its efficiency in terms of latency and its robustness in different settings.
\end{abstract}





\maketitle

\begin{sloppypar}
\vspace{-0.2cm}
\section{Introduction}
Large Language Models (LLMs) have demonstrated impressive capabilities in generating coherent and contextually relevant responses to their prompts, leading to their widespread adoption in various natural language understanding and generation tasks \cite{xu2024searchinthechaininteractivelyenhancinglarge,ram-etal-2023-context}. However, their rise has also brought substantial trustworthiness challenges, raising concerns of correctness and factuality of the generated text, which is referred to in the literature as \textit{hallucination} \cite{bechard2024reducing, shuster2021retrieval}.  One prominent solution to mitigate hallucination is Retrieval-Augmented Generation (RAG), a framework that enhances the performance of LLMs by prepending to the prompt a context composed of background documents \cite{dao2023chatgptgoodbingchat, izacard2022atlasfewshotlearningretrieval}. However, recent studies reveal that  hallucination remains a persistent challenge in RAG models \cite{krishna-etal-2021-hurdles} since they still generate incorrect responses and, more importantly, lack \textit{faithfulness} \cite{Qi_2024,wallat2024correctness,song2025measuringenhancingtrustworthinessllms}. This is while ensuring not only correct  but grounded responses with accurate document sources, is  particularly essential in high-stakes domains such as healthcare \cite{li2024mitigating}, legal reasoning \cite{magesh2024hallucination}, and more generally in information-decision making. \\
\begin{figure}
    \centering
    \includegraphics[width=1.02\linewidth]{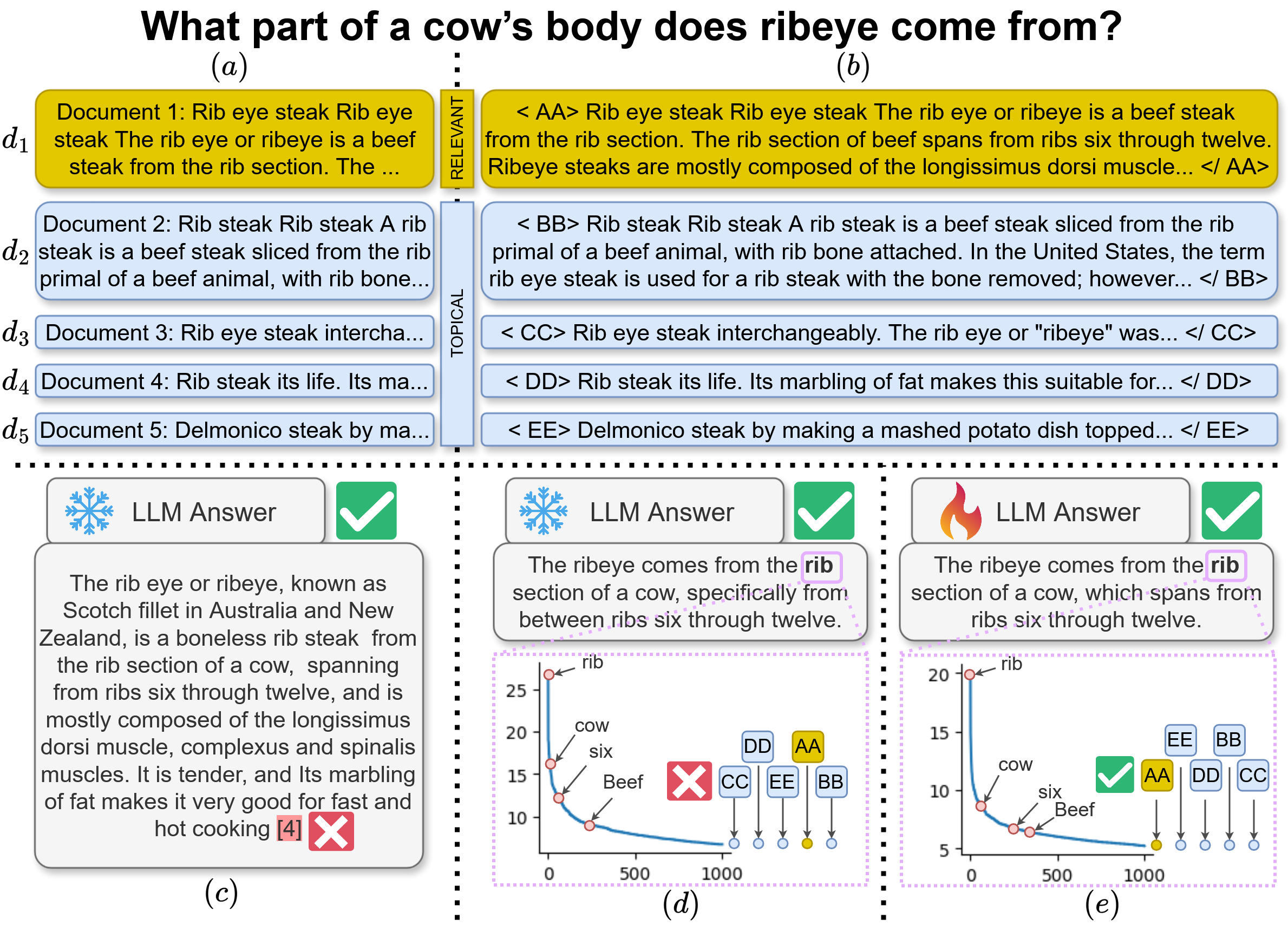}
    \vspace{-0.9cm}
    \caption{(a) and (c) depict the input and output of a naive RAG model with citations. (d) and (e) show the decreasing logits distribution when generating the token ``rib'' of a (d) frozen LLM, and a (e) finetuned LLM, \textbf{LoDIT}, when using our marking proposal presented in (b).}
    \label{fig:motivationtolenatt}
    \vspace{-0.5cm}
\end{figure}
To address this issue, recent research has turned to document \textit{attribution}, which aims to link model-generated answers with citations to specific retrieved documents. 
Yet, existing attribution methods commonly suffer from major limitations. Self-generated methods rely on the LLMs' capabilities to generate an answer and \textit{citations} jointly \cite{song2025measuringenhancingtrustworthinessllms, huang2024learningfinegrainedgroundedcitations, xia2024groundsentenceimprovingretrievalaugmented, menick2022teachinglanguagemodelssupport}, leading to citations most of the times treated as a post-hoc process which relies on the model to retrospectively justify its outputs rather than faithfully reflect its reasoning process. Retrieval-based methods leverage external sources of information to produce post-hoc attributions \cite{huang2024advancinglargelanguagemodel, slobodkin-etal-2024-attribute, gao-etal-2023-rarr, sancheti-etal-2024-post}. Similarly, these methods tend to identify plausible rather than faithful sources, weakening the reliability of the attribution \cite{wallat2024correctness,song2025measuringenhancingtrustworthinessllms}. Recently, \textit{models-internals} methods have tackled this issue by using open-LLMs and leveraging internal model signals, such as attention weights or gradient-based measures, to induce attributions that are more interpretable and better aligned with the model’s actual decision-making process \cite{ding2024attentiondependencyparsingaugmentation, Qi_2024, cohenwang2024contextciteattributingmodelgeneration, phukan-etal-2024-peering}. 
However, these methods either do not explicitly capture the causal influence between each retrieved contextual document and the generated output \cite{phukan-etal-2024-peering,cohenwang2024contextciteattributingmodelgeneration} or approximate it based on  attributions estimated at the whole context level \cite{Qi_2024}, which is prone to inaccuracy, particularly in the case of long contexts. Furthermore, these approaches induce additional latency by estimating answer attributability through multiple model generation passes with and without context.




This paper addresses these issues by proposing a method for model-intern based attributed answer generation  which complies with two key desiderata: (1) \textit{explicit attribution}: to better favour faithfulness, the model should  provide direct evidence about the causal relationship between its outputs and each document in the context; (2) \textit{efficiency: }: low latency of model deployment by gaining the ability to jointly generating and attributing answers.  Specifically, this paper investigates the following research question: “\textit{Can we leverage LLM's predictions through token logits to jointly generate the answer to a query and  ground it with documents in a  context?}”. This question gives rise to another underlying question of whether the model's output token logit could be representative of the content of an input context document. We build our thoughts on recent work related to setwise prompting strategies for document relevance estimation \cite{Zhuang_2024}. The major finding is that, given a query and a set of input document passages prefixed with token identifiers, as in multi-choice question answering (MCQA), the model output document-identifier logit is a good estimation of passage relevance to the query, leading to effective document ranking. Thus, we could reasonably conclude that the token-identifier logit is a good rationale to estimate the contribution of the associated document content to the generation process of the LLM. Relevant to our work, we investigate to what extent the token-identifier logit associated with an input document context might be indicative of the model's reasoning process during answer generation. However, recent work has pointed out that LLMs suffer from the selection bias \cite{wei-etal-2024-unveiling, zheng2024large,10.1145/3637528.3671458} and token-bias \cite{jiang2024peektokenbiaslarge}.  The LLMs' output token logits do not entirely depend on the context but are internally deviated mostly due to the distribution of their pre-training.  \\
In light of these considerations, we propose \textbf{LoDIT} (\textbf{Lo}gits of \textbf{D}ocument \textbf{I}dentifier \textbf{T}okens), a method for attributed retrieval-augmented answer generation that emphasizes \textit{faithful} generation induced by \textit{debiaised} predicted model token logits. 
A motivation example of our proposal is presented in Figure \ref{fig:motivationtolenatt}. The input and output of a \textit{naive RAG model with citations} is depicted in (a) and (c), with the LLM giving the right answer to the query while generating an inaccurate citation. 
Answers from a vanilla LLM (d) and \textbf{LoDIT} (e) are presented for the marked context (b). Note that both LLMs also provide the correct answer, and a focus is made on the step that generates the first occurrence of the token ``rib'' in both answers (plotted values correspond to top 1000 observed logits in decreasing order and highlight a set of selected tokens). The relative order of logits values of the generated word and related words (``cow'', ``six'', and ``Beef'') is the same but not the values for the document identifiers (`` AA'', `` BB'', etc.) where \textbf{LoDIT} is able to rank higher (but not in top 1000) the correct document to be attributed. 

\textbf{LoDIT} consists of two main stages. We refer to the first stage as  \textit{token-level contribution} estimation, where context documents in the prompt are first marked with token identifiers.  We explore several marking strategies, associating identifiers with documents. Then, the LLM' generated answer tokens are \textit{explicitly} attributed to context documents based on \textit{debiaised} document-identifier token logits using a joint fine-tuning on answer generation and attribution. Each answer token is assigned a set of scores that estimate its attributability to each document  with little additional latency. This stage is followed by a \textit{statement-level attribution} based on a top-k pooling aggregation function to favour high-level faithfulness of the attribution. We evaluate \textbf{LoDIT} under the Trust-Align benchmark \cite{song2025measuringenhancingtrustworthinessllms}, which promotes trustworthiness in RAG. 
The results show that \textbf{LoDIT} significantly improves the performance upon both the state-of-the-art faithful attribution methods and trustworthiness-focused attribution generation methods. The main contributions of this paper can be summarized as: 
\begin{enumerate}
    \item We propose a novel  method for faithful attributed answer generation in RAG. We show that the logits of document-identifier tokens associated with retrieved documents in the context can explicitly capture the causal effect of context on answer generation. 
    \item We introduce a joint fine-tuning strategy to explicitly and directly learn to attribute generated answer tokens to documents in the context,  mitigating the selection and token bias in LLMs. 
    \item We perform comprehensive experiments on three standard datasets, ELI5, ASQA, and QAMPARI, using the recent Trust-Align Framework \cite{song2025measuringenhancingtrustworthinessllms} better aligned with faithfulness. The results validate both the effectiveness and efficiency of \textbf{LoDIT}. Our code is publicly available \footnote{available after notification}. 
\end{enumerate}
\vspace{-0.2cm}
\section{Related Work}

\subsection{Attributed answer generation}\label{sec:attributionmethods}

Existing attribution methods fall into three categories \cite{ding2024attentiondependencyparsingaugmentation}. Self-generated methods consider attribution as a generative process. They jointly generate answers and attributions for these answers \cite{song2025measuringenhancingtrustworthinessllms, huang2024learningfinegrainedgroundedcitations, xia2024groundsentenceimprovingretrievalaugmented, menick2022teachinglanguagemodelssupport}. For instance, \textit{FRONT} \cite{huang2024learningfinegrainedgroundedcitations} fine-tunes an LLM with a fine-grained attribution framework composed of two steps: selecting supporting quotes from documents and using these quotes to guide generation. These methods have the advantage of harnessing the powerful capabilities of LLMs as well as requiring no additional setup to obtain attributions. However, these methods lack interpretability and faithfulness due to their inherent black-box nature. \\
Retrieval-based methods deliver sentence-level citations by extracting relevant information from external sources, ensuring that each sentence is properly cited \cite{huang2024advancinglargelanguagemodel, slobodkin-etal-2024-attribute, gao-etal-2023-rarr, sancheti-etal-2024-post}.
For example, START \cite{huang2024advancinglargelanguagemodel} fine-tunes an LLM to first generate an answer and then obtain attribution with several steps: decomposing the answer into multiple claims, combining similar claims, and finally generating a synthetic document that covers those claims. These strategies are  more interpretable than self-generated black-box LLMs, but introduce additional latency and suffer similarly as the self-generated methods, of the lack of faithfulness of the obtained \textit{attributions}.\\
Close to our work, models-internals methods leverage statistics and similarity metrics about internal components of LLM models to align the generated answer with the prompt, thus inducing attribution \cite{ding2024attentiondependencyparsingaugmentation, Qi_2024, cohenwang2024contextciteattributingmodelgeneration, phukan-etal-2024-peering}.
For instance, \textit{MIRAGE} \cite{Qi_2024} first identifies \textit{sensitive} tokens in the answer by computing the shift in the LLM predictive distribution when removing the documents from the prompt. Then, they perform a finer-grained analysis using \textit{contrastive feature attribution}  to estimate which part of the prompt is more impactful on the generation. Ding et al. \cite{ding2024attentiondependencyparsingaugmentation} propose a dependency parsing method that first recognizes key answer tokens within atomic facts in the context and then assigns attribution scores in context by aggregating attention weights between the response and the prompt.  
These methods offer higher interpretability and greater control over the attribution granularity. Furthermore, by directly leveraging statistics from the model, the obtained \textit{attributions} tend to be considered more faithful.\\
However, these methods rely on indirect and two-level based causal relationships between tokens and contexts, thereby inducing additional latency to measure attributions. In contrast, \textbf{LoDIT} relies on a direct and simple token-level, yet effective, attribution method that leverages internal causal statistics from the generative model, token by token. It jointly performs generation as well as faithful causal attribution, leading to little added latency during inference. Moreover, we couple our novel attribution method with an evaluation framework specifically built for trustworthiness in RAG, yielding better insights about the faithfulness of our proposed method.
\vspace{-0.5cm}
\subsection{Evaluation of attributed generation}\label{sec:attributionevaluation}
The task of attributed answer generation is traditionally evaluated along two axes: answer correctness and attribution quality. Answer correctness relies on traditional measures used for the evaluation of  semantic-based metrics including ROUGE \cite{lin2004rouge}, BLEU \cite{papineni2002bleu}, and BertScore \cite{zhang2019bertscore}. The evaluation of attribution models has given rise to the design of new metrics  aligned with model capacity to  output documents supporting the generated answer \cite{liu-etal-2023,yue-etal-2023-automatic,gao2023enablinglargelanguagemodels,gao-etal-2023-rarr,DjeddalETSPKT24}. For instance, Gao et al. proposed the framework ALCE \cite{gao2023enablinglargelanguagemodels} using the citation recall and citation precision metrics by relying on the automatic binary score of an  NLI classifier between the statements and the citations in the attribution. In Yu et al. \cite{yue-etal-2023-automatic}, the authors propose the ATTscore, which evaluates specific dimensions of the attribution, such as  the exploratory dimension, which covers the comprehensiveness rate of the attribution and the contradiction. 
Recently, Song et al. \cite{song2025measuringenhancingtrustworthinessllms}, introduced the \textit{Trust-Align} work evaluation framework. The authors have shown the limitations of these traditional metrics when considering trustworthiness in RAG. 
They subsequently refine the traditional metrics into trustworthiness-oriented measures and propose a training framework directly focused on improving the overall trustworthiness of a RAG framework, including faithfulness. 
Our work leverages the \textit{Trust-Align} evaluation framework with the objective of building a trustworthy attributed generation model yielding faithful attributions. 

\vspace{-0.5cm}
\subsection{Selection biais of LLMs}\label{sec:llmbiais}
During a generation process, an LLM leverages a multitude of factors to compute a specific token's logit (and generation probability): the input tokens, their positions, as well as internal knowledge assimilated during training.
Recent work has highlighted the selection bias \cite{wei-etal-2024-unveiling, zheng2024large, zheng2023judgingllmasajudgemtbenchchatbot, 10.1145/3637528.3671458} and token-bias \cite{jiang2024peektokenbiaslarge} of LLMs, showing that the probability of generating a given token does not entirely depend on the generation context, but on biases internalized by the model during training.
This bias is especially strong in selection tasks where the LLM has to select an identifier based on the relevance of its assigned piece of text (e.g., MCQA), with several works showing that LLMs inherently favor some identifiers over others \cite{wei-etal-2024-unveiling, zheng2024large, zheng2023judgingllmasajudgemtbenchchatbot, 10.1145/3637528.3671458}.
To reduce these biases, Wei et al. \cite{wei-etal-2024-unveiling} propose calibrating the LLMs' probabilities through multiple inferences of multiple token permutations based on a fixed dataset.
\vspace{-0.2cm}
\section{Problem Statement}
\vspace{-0.1cm}
\subsection{Preliminaries and notations}\label{sec:prelim}
\paragraph{\textbf{Attribution}} 
Let us consider  an LLM $\mathcal{M}$ and $S$ a statement generated by $\mathcal{M}$.
An attribution $a$ of statement $S$ is composed of a set of document identifiers  $\{id_i\}_{i=1}^{n_S}$  citing document passages $\{d_i\}_{i=1}^{n_S}$ to support or ground statement $S$. 
An attribution $a$ can be either model-generated or not. An attributed answer statement $S^{a}$ is composed of the concatenation of a model-generated statement $S$ and its attribution $a$. 
%
\vspace{-0.2cm}
\paragraph{\textbf{Faithful attribution}}
  Let us consider  an LLM $\mathcal{M}$, $C$ a context composed of reference documents $C =\{d_i\}_{i=1}^{K}$,  $S$ a statement generated by $\mathcal{M}$ and $a =\{id_i\}_{i=1}^{n_S}$ an attribution of $S$.
  By adopting the definition of faithfulness reported in recent work \cite{wallat2024correctness,song2025measuringenhancingtrustworthinessllms}, we assess $S^{a}$ as \textit{faithful} if it satisfies the following conditions with increasing restriction levels: (1) \textit{context-relatedness}: each document-identifier in the attribution cites a document in the context, i.e., $\forall id_i \in a $, $d_i\in C$; (2) \textit{groundedness}: $S$ is factually supported by the content of the documents cited in the attribution $\{d_i \vert id_i \in a\}$; (3) \textit{causal model-generation}: to generate the statement $S$, model $\mathcal{M}$ must rely causally on the tokens of documents cited in the attribution $a$.

\vspace{-0.3cm}
\subsection{Task definition}
 In this work, we focus on retrieval-augmented attributed answer generation.
 The goal of this task  is to answer query $q$ with  answer $A$ using a retrieved context $C$ from a corpus of documents $D$, $C=\{d_1\dots d_K \vert d_i \in D\}$. Answer $A$ is composed of a set of attributed answer statements $\{S_1^a\ldots S_{m}^a\}$ where $S_i^a$ is the concatenation of the answer statement $S_i$ and each element of its attribution $a_i=\{id_{k} \in T\}_{k=1}^{k_i}$. Consider an LLM $\mathcal{M}$ pre-trained over a tokenizer dictionary $T$,  instructed with a  query $q$,  and context $C$.
The main problem we tackle in our work  consists of inducing attributions $\{a_i\}_{i=1}^m$ to  answer statements $\{S_i\}_{i=1}^m$ generated by $\mathcal{M}$ such as $\{S^a_i\}_{i=1}^m$ are faithful. Beyond \textit{context-relatedness} and \textit{groundedness}, we specifically address in our work, faithfulness in terms of \textit{causal model generation}. 

\begin{figure}
    \centering
    \includegraphics[width=1.05\columnwidth]{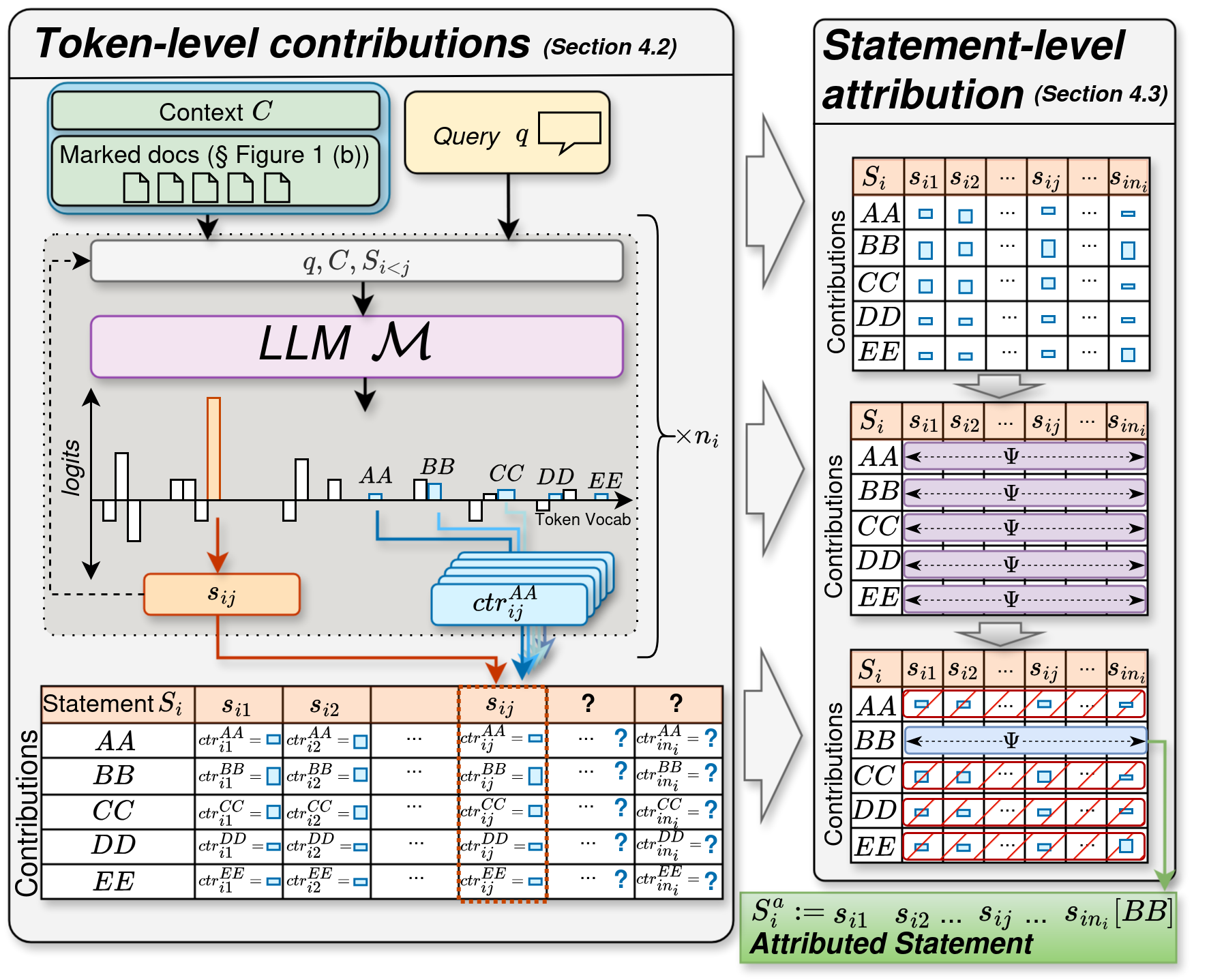}
    \vspace{-0.5cm}
    
    \caption{Overview of our proposal, \textbf{LoDIT}. Given a query and a context composed of documents, \textbf{LoDIT} jointly generates an answer and induces faithful causal attributions, grounding the answer into the context. The first step (left) consists in computing token-level contributions of marked documents with specific tokens and collecting the logits associated with these tokens during generation. Once a statement has been generated, a second step (right) aggregates the computed contributions into statement-level attribution. This illustration features the $marking_{BA}$ strategy.}\label{fig:overviewfig}
\vspace{-1.3cm}
\end{figure}
\vspace{-0.3cm}
\section{Logit-induced Answer Generation and Attribution}\label{sec:computingcontrib}
\vspace{-0.1cm}
\subsection{Method overview}




The overview of \textbf{LoDIT} is shown in Figure \ref{fig:overviewfig}. We assume access to a generative LLM $\mathcal{M}$ pre-trained over a tokenizer dictionary $V=\{\tau_{1}, \tau_{2},\dots,\tau_{\vert V\vert}\}$. Given a query $q$ and a context $C$ retrieved from a corpus of documents $D$, $C=\{d_1\ldots d_K \vert d_i \in D\}$, $\mathcal{M}$ generates an output answer as a set of statements $A=\{S_1\ldots S_{m}\}$. Each statement $S_i$ is the concatenation of tokens $ s_{ij} \in V$, with $S_i(s_{i1}\circ\dots \circ s_{in_i})$. $\mathcal{M}$ continuously generates  the tokens of answer statement $S_i$, $tok(S_i)\subset V$, by sampling the next token $s_{ij}$ based on the probability distribution 
$P(V\vert q, C,A_{S_{<i}},S_{i<j},\mathcal{M})=\{p(s_{ij}\vert q,C,A_{S_{<i}},S_{i<j},\mathcal{M})\}_{i=1}^{\vert V\vert}$, where $A_{S_{<i}}=\{S_1 \dots S_{i-1}\}$ are the previously generated answer statements, $S_{i<j}$ are the previous generated tokens of $S_i$,
and $\mathcal{M}(s_{ij} \vert q, A_{S_{<i}},S_{i<j},C)$ is the predicted \textit{token logit} of $s_{ij}$ before the softmax that transforms logit scores into probabilities, such as: $p(s_{ij}\vert q, C, A_{S_{<i}},S_{i<j})=\frac{exp(\mathcal{M}(s_{ij}\vert q,C,A_{S_{<i}},S_{i<j}))}{\sum_{j=1}^{\vert V\vert} exp(\mathcal{M}(s_{ij}\vert q, C,A_{S_{<i}},S_{i<j}))}$.\\

To  ensure faithful attribution in terms of causal model generation, we leverage previous findings showing the potential of document-identifier token logits to indicate the relevance of associated document content \cite{Zhuang_2024}. The authors have shown that the  LLM label outputs logits can be used to estimate the likelihood of document content relevance to a query based on labeled documents fed to the LLM in the context. Grounded on this result, we postulate that the logits of label tokens could bridge between source documents in the input context and  their representative labels in the LLM output. Thus, we  use identifier tokens to mark the input documents and then leverage their logits to provide clues on their relevance to support LLM answer generation, token by token.  \\
In light of this insight,  we first mark each document passage $d_i\in C$ with a document identifier $id_i\in V$, resulting in documents $md_i$ and context $C=\{md_1\ldots md_K \}$.
At each generation step of answer tokens $s_{ij}$, the logit of each token-identifier $id_k$, noted $l_k$, is used as the basis of \textit{token-level score contribution}, noted $ctr_{ij}^k$, of document $d_k\in C$ to generate answer token $s_{ij}$.  To mitigate the token bias and position bias of LLMs \cite{wei-etal-2024-unveiling,zheng2024large}, we fine-tune $\mathcal{M}$ through a co-training on the answer generation task and a logit-based debiaising attribution task  such as $ctr_{ij}^k=\mathcal{M}_{aa}(id_k\vert A_{S_{<i}},S_{i<j},C)$, with $M_{aa}$ being the optimized debiaised model. We further aggregate the token-level contribution scores to induce the statement-level attribution score using an aggregation function  $\Psi$, such as  $a_i=\Psi\left[\{ctr_{ij}^k\}_{j=1}^{\vert n_i\vert}\right]_{k=1}^{K}$. 
\vspace{-0.1 cm}
\subsection{Learning token-level contribution using LLM's logits} 

\subsubsection{Document marking}\label{sec:marking}
\textit{Marking} functions are one of the core components of \textbf{LoDIT}. A \textit{marking} function is a function that prefixes  the content of documents in context $C$ with identifier tokens. We investigate three \textit{marking} functions, presented in Table \ref{tab:markingstrats}. Each function transforms each document $d$ in the context $C$ into a marked document $md_i$ agnostically to the nature of the identifier token $id_i$:
\begin{itemize}
\vspace{-0.1 cm}
    \item \colorbox{LightCyan}{$md_{i} = marking_{BA}(d_i, id_i)$}: we incorporate the identifier token $id_i$ \textbf{B}efore and \textbf{A}fter the document $d_i$ using greater-than $(>)$ and less-than $(<)$ signs. The objective of this strategy is to leverage the LLM's capabilities of processing code-like structure, motivated by the huge amount of code data present in LLM's training phases. 
    \item \colorbox{LightCyan}{$md_{i} = marking_{BAS}(d_i, id_i)$}: we incorporate the identifier token $id_i$ \textbf{B}efore and \textbf{A}fter each \textbf{S}entence of the document $d_i$ using greater-than and less-than signs. This is a similar yet finer-grained strategy than the one above.
    \item \colorbox{LightCyan}{$md_{i} = marking_{AW}(d_i, id_i)$}: we incorporate the identifier token $id_i$ before \textbf{A}ll \textbf{W}ords of document $d_i$. The objective of this strategy is to study whether repetition allows the LLM to yield higher logits for the identifier tokens during generation. \\
\end{itemize}
\vspace{-0.5cm}



\subsubsection{Debiaising LLM' token logits to learn document contribution}\label{sec:finetuningLogits}\hfill
\begin{table*}[]

\resizebox{\textwidth}{!}{%
\begin{tabular}{c|l}
\textbf{Marking function} & \multicolumn{1}{c}{\textbf{Marked documents}}                                                                                                                                                                                                                                                                                                             \\ \hline
$marking_{BA}$               & \begin{tabular}[c]{@{}l@{}}\textless{} \textcolor{teal}{\textit{AA}}\textgreater{}The rib eye or ribeye is a beef steak from the rib section. The rib section of beef spans from ribs six through twelve.\textless{}/ \textcolor{teal}{\textit{AA}}\textgreater\\ \textless{} \textcolor{purple}{\textit{BB}}\textgreater{}A rib steak is a beef steak sliced from the rib primal of a beef animal.\textless{}/ \textcolor{purple}{\textit{BB}}\textgreater{}\end{tabular}                                                     \\ \hline
$marking_{BAS}$     & \begin{tabular}[c]{@{}l@{}}\textless{} \textcolor{teal}{\textit{AA}}\textgreater{}The rib eye or ribeye is a beef steak from the rib section.\textless{}/ \textcolor{teal}{\textit{AA}}\textgreater \textless{} \textcolor{teal}{\textit{AA}}\textgreater{}The rib section of beef spans from ribs six through twelve.\textless{}/ \textcolor{teal}{\textit{AA}}\textgreater\\ \textless{} \textcolor{purple}{\textit{BB}}\textgreater{}A rib steak is a beef steak sliced from the rib primal of a beef animal.\textless{}/ \textcolor{purple}{\textit{BB}}\textgreater{}\end{tabular} \\ \hline
$marking_{AW}$               & \begin{tabular}[c]{@{}l@{}}\textcolor{teal}{\textit{AA}}The \textcolor{teal}{\textit{AA}} rib \textcolor{teal}{\textit{AA}} eye \textcolor{teal}{\textit{AA}} or \textcolor{teal}{\textit{AA}} ribeye \textcolor{teal}{\textit{AA}} is \textcolor{teal}{\textit{AA}} a beef \textcolor{teal}{\textit{AA}} steak \textcolor{teal}{\textit{AA}} from [...] \textcolor{teal}{\textit{AA}} spans \textcolor{teal}{\textit{AA}} from ribs \textcolor{teal}{\textit{AA}} six \textcolor{teal}{\textit{AA}} through \textcolor{teal}{\textit{AA}} twelve.
\\  \textcolor{purple}{\textit{BB}} A \textcolor{purple}{\textit{BB}} rib \textcolor{purple}{\textit{BB}} steak \textcolor{purple}{\textit{BB}} is \textcolor{purple}{\textit{BB}} a \textcolor{purple}{\textit{BB}} beef \textcolor{purple}{\textit{BB}} steak \textcolor{purple}{\textit{BB}} sliced \textcolor{purple}{\textit{BB}} from \textcolor{purple}{\textit{BB}} the \textcolor{purple}{\textit{BB}} rib \textcolor{purple}{\textit{BB}} primal \textcolor{purple}{\textit{BB}} of \textcolor{purple}{\textit{BB}} a \textcolor{purple}{\textit{BB}} beef \textcolor{purple}{\textit{BB}} animal.\end{tabular}                                                                                                                                               

\end{tabular}
}
\caption{Illustration of the marking strategies. Two documents are marked with identifier tokens ``\textcolor{teal}{\textit{AA}}'' and ``\textcolor{purple}{\textit{BB}}''.}\label{tab:markingstrats}
\vspace{-0.7 cm}
\end{table*}
\paragraph{\textbf{Motivation}}
Our key idea is to induce token contribution to answer attribution along  the LLM answer generation by leveraging document-identifier token logits.  However, many previous works have pointed out issues about LLM bias that would hinder the performance of the task \cite{zheng2024large,wei-etal-2024-unveiling}:  (1) \textit{selection bias}:  LLMs are vulnerable to option positions in the context. Regarding our task, this would make
document-identifier tokens' logits biased by the rank of the documents in the context instead of leveraging their content to support the answer; (2) \textit{token bias}: LLMs assign more probabilistic mass to specific tokens based on the distribution of pre-training data. This bias could lead to overestimating resp. underestimating document-identifier logits because of their high frequency vs. low frequency in the pre-training stage. This would critically impact the final attribution since we leverage the logit of the document-identifier token to estimate the contribution of its associated document, especially as the corresponding probabilities get substantially low.  One simple way to mitigate these biases is to ablate parts of the prompt (e.g., documents) and use their impact on the generation probability distribution $P$ as an indication of the attribution, as done in previous work  \cite{Qi_2024, cohenwang2024contextciteattributingmodelgeneration}. The major drawback of this ablation and repeat approach is the necessity of performing two LLM generation processes (one with document and one without), which introduces additional latency, as shown in our analyses (§ \ref{sec:Effectsoflogitdebiaising}).\\
In addition to selection and token bias, one particularity of our problem is that we  aim to debias only a few logits document-identifier tokens  in comparison to the token vocabulary size of the LLM $(K<<\vert T\vert)$, while keeping the others reflecting the LLM's certainty about the next answer token to be predicted.  To tackle all the aforementioned issues, we propose fine-tuning a backbone LLM $\mathcal{M}$ jointly guided by  the answer generation task and an attribution task.



\paragraph{\textbf{Document contribution through logits learning}}

  To mitigate the issues mentioned above, we propose fine-tuning the backbone LLM  $\mathcal{M}$ jointly on the tasks of answer generation and attribution using a training dataset $\mathcal{D}_{train}$ wich consists in a set of gold triplets $(q,C,\hat{A})$ with $\hat{A}$ the gold attributed answer of query $q$ given context $C$, such as $\hat{A} = \{\hat{S}_1^a\ldots \hat{S}_{m}^a\}$ where $\hat{S}_i^a$ is the concatenation of the gold answer statement $\hat{S}_i$ and each element of its attribution $\hat{a}_i=\{id_{k} \in T \vert d_k\in C \}_{k=1}^{k_i}$. Specifically, $\mathcal{M}$ learns to fit the logits of associated document-identifier tokens to achieve answer correctness and attribution faithfulness.  
The optimization problem is defined as the minimization of the attributed answer generation loss $\mathcal{L}_{aa}$, which is a combination of the answer loss $\mathcal{L}_{ans}$ and attribution loss $\mathcal{L}_{att}$: 
\begin{equation}
\begin{split} 
\label{eq:mselogits}
\mathcal{L}_{aa} = \sum_{(q, C, \hat{A}) \in \mathcal{D}_{train}} \sum_{\hat{S}_i \in \hat{A}}(1 - \alpha) \mathcal{L}_{ans} + \alpha \mathcal{L}_{att} 
\end{split} 
\end{equation}
where $\alpha$ is a weight balancing the answer and attribution losses. The answer loss $\mathcal{L}_{ans}$ is the basic LLM next token prediction cross-entropy loss:

\begin{equation}
\mathcal{L}_{ans} = \sum_{ \hat{s}_{ij}\in tok(\hat{S}_i)} p_{ij} log (p_{ij})
\end{equation}

where $p_{ij}=p(\hat{s}_{ij}\vert q, \hat{A}_{\hat{S}_{<i}},S_{i<j},C,\mathcal{M})$.\\

The attribution loss is an MSE loss, which has shown to be effective for learning specific logits \cite{jiao2020tinybertdistillingbertnatural, ijcai2021p362}, computed solely for the document identifier tokens $\{id_k\}_{k=1}^{k_i}$. $\mathcal{L}_{att}$ is defined as: \\
\begin{equation}
\mathcal{L}_{att} = \sum_{ \hat{s}_{ij}\in tok(\hat{S}_i)} \sum_{ id_k\in \hat{a}_i} (l_k-\hat{l}_k)^2
\end{equation}

where $l_k=\mathcal{M}(id_{k}\vert q, \hat{A}_{S_{<i}},S_{i<j},C)$ and $\hat{l}_k$ are labels focusing solely on the logits to predict for document-identifier tokens $\{id_k\}_{k=1}^{k_i}$.
We use scaled logits labels $\hat{l}_k$  with values aligned with the decreasing conditions of groundedness  and context-relatedness conditions of answer faithfulness (§ \ref{sec:prelim}). We empirically set up these labels as described in § \ref{sec:taininglabels}.

The contribution $ctr_{ij}^k$ of document context $d_k$ to the generation of answer token $s_{ij}$ is computed at inference based on the logit output of the fine-tuned LLM model $\mathcal{M}_{aa}$ for the token-identifier $id_k$, such as:
\begin{equation}\label{eq:ctr}
ctr_{ij}^k=\mathcal{M}_{aa}(id_{k}\vert q, A_{S_{<i}},S_{i<j},C)
\end{equation}

The contribution $ctr_{ij}^k$ captures an explicit causal influence between model token generation and document in context, leading to interpretable and smoothly aligned attribution with the model’s actual decision-making process.

\subsection{From token-level to statement-level answer attribution} 
\label{sec:aggreg}

The objective here is to induce for each model-generated statement $S_i(s_{i1}\circ\dots \circ s_{in_i})$ an attribution $a_i=\{id_k\}_{k=1}^{k_i}$ composed of document-identifier tokens. To this end, we mainly rely on  the token-level contributions $ctr_{ij}^k$ computed at each generated token $s_{ij}$, which we aggregate using an aggregation function $\Psi$, such as $a_i=\Psi\left[\{ctr_{ij}^k\}_{j=1}^{\vert n_i\vert}\right]_{k=1}^{K}$. The contribution scores $ctr_{ij}^k$  are inherently non-discrete, allowing for a finer understanding of the attribution mechanism and improving interpretability. However, user-oriented systems benefit from having discrete attributions as it is less tedious to process for a human.
To be more aligned with this requirement, we use a contribution aggregation function $\Psi$, which transforms the token-level contributions into boolean attribution, applicable to each document $d_k$ in the context $C$ and returning $1$ if $d_k$ is to be included in $a_i$, $0$, otherwise. Specifically, we define the following aggregation function:

\begin{equation}\label{eq:aggrg}
a_i = \{id_k \vert \sum_{j=1}^{n_i}   \mathbf{1}_{ctr_{ij}^k > \phi_{\text{prop}} }>   \lambda (n_i)\}_{k=1}^{K}
\end{equation}

where $\mathbf{1}_{ctr_{ij}^k > \phi_{\text{prop}}= 1}$ if $ctr_{ij}^k > \phi_{\text{prop}}$ and 0 otherwise. $\phi_{prop}$ is a contribution threshold and $\lambda$ is a token proportion threshold (e.g., a percentage). Similar to a top-k pooling \cite{kalchbrenner2014convolutionalneuralnetworkmodelling}, the intuition behind this aggregation operator is that documents should be attributed to the answer statement if they contribute highly to at least a thresholded percentage of the associated tokens. 
\\
\vspace{-0.5cm}
\section{Experimental setup}

\subsection{Trustworthiness-focused evaluation benchmark}

Our evaluation is grounded in recent studies on model trustworthiness in RAG, the \textit{Trust-Align} framework \cite{song2025measuringenhancingtrustworthinessllms}, which builds upon the ALCE benchmark \cite{gao2023enablinglargelanguagemodels} by refining its evaluation metrics so that they not only measure the capabilities of models to answer correctly and provide accurate citations, but also refuse to answer when not relevant information is provided by the retrieved documents.

The ALCE benchmark is an attributed text-generation benchmark that covers three long-form Question Answering (QA) datasets, spanning various types of question: \textbf{QAMPARI} \cite{amouyal2023qampariopendomainquestionanswering} is a factoid QA dataset based on Wikipedia, with answers presented as a collection of entities; \textbf{ASQA} \cite{stelmakh-etal-2022-asqa} is a long-form factoid QA dataset featuring inherently ambiguous questions that necessitate multiple concise answers to represent diverse perspectives; \textbf{ELI5} \cite{fan-etal-2019-eli5} consists of open-ended questions designed to be simplified for the understanding of a five-year-old, requiring explanatory responses spanning multiple sentences. 

We evaluate the effectiveness of \textbf{LoDIT} on each dataset using the metrics proposed in the \textit{Trust-Align} framework \cite{song2025measuringenhancingtrustworthinessllms}: 
\vspace{-0.1cm}
\begin{itemize}
    \item \textbf{F1AC} which measures the answer correctness by specifically considering questions that can be answered using the documents in context, i.e., questions with enough relevant information in the documents.
    \item \textbf{F1GR}, which measures the capabilities of the model to refuse to answer a question when it cannot be answered using the retrieved document.
    \item \textbf{F1GC}, which measures citation groundedness: the capabilities of the model to output grounded citations to its answers. This metric evaluates the \textit{context-relatedness} and \textit{groundedness} conditions of attribution faithfulness defined in § \ref{sec:prelim}.
    \item \textbf{TRUST}, which averages the three former metrics into a single trustworthiness score.
\end{itemize}
\vspace{-0.1cm}
We refer the reader to the \textit{Trust-Align} paper \cite{song2025measuringenhancingtrustworthinessllms} for the full motivation and thorough explanation of these metrics.
\vspace{-0.2cm}
\subsection{Baselines}

We evaluate our approach, \textbf{LoDIT}, as well as zero-shot prompting a Llama3.1 8B with each of our marking strategies for completeness.

We compare our result against several baselines:
\begin{itemize}
  \item{4 baselines whose results are reported from \cite{song2025measuringenhancingtrustworthinessllms}:}
    \begin{itemize}
    \item Llama3 8B with \textit{PostCite}* \cite{gao2023enablinglargelanguagemodels}, which first generates answers and then retrieves documents to perform attribution.
    \item Llama3 8B with \textit{PostAttr}* \cite{gao2023enablinglargelanguagemodels}, which first generates answers and then uses an entailment language model (TRUE-NLI) to perform attribution.
    \item Llama3 8B fine-tuned with \textit{FRONT}* \cite{huang2024learningfinegrainedgroundedcitations}, which leverages fine-grained attributions to improve citations grounding.
    \item Llama3 8B fine-tuned with \textit{Trust-Align}* \cite{song2025measuringenhancingtrustworthinessllms}, which is specifically designed to improve models' trustworthiness.
    \end{itemize}
  \item{\textit{MIRAGE} \cite{Qi_2024} with Llama 3.1 8B, which identifies \textit{sensitive} tokens in the answer and then computes the contribution of each document to these sensitive tokens by computing statistics with and without parts of the prompt. The official implementation with default parameters has been used. We included this baseline as it is a faithful attribution method close to our work, but no results on the \textit{Trust-Align} benchmark were available.}
  \item{The Llama3.1 8B model:}
   (1) with Zero-Shot prompting, using the same prompt as in \cite{song2025measuringenhancingtrustworthinessllms}; (2) with fine-tuning on the \textit{Trust-Align} training dataset, to ensure fair comparison by training on the same LLM as well as to perform statistical significance tests between models.
\end{itemize}

\subsection{Training datasets}

Our training data is composed of three sources: Trust-Align \cite{song2025measuringenhancingtrustworthinessllms}, Hagrid \cite{kamalloo2023hagridhumanllmcollaborativedataset} and ExpertsQA \cite{malaviya2024expertqaexpertcuratedquestionsattributed}.

\begin{itemize}
    \item The \textbf{Trust-Align} dataset focuses on improving models' trustworthiness by building upon the  QAMPARI, ASQA, and ELI5 training sets.
One of the features of this dataset is that it is focused on making the models more robust to hallucination by teaching them to refuse to answer when in the absence of relevant information in the documents.
These examples amount to approximately 25\% of the training set.

\item \textbf{Hagrid} \cite{kamalloo2023hagridhumanllmcollaborativedataset} is a generative information retrieval dataset constructed on top of the MIRACL dataset \cite{10.1162/tacl_a_00595}. It contains human-based annotations of LLM-generated answers.
We preprocess the data by only selecting examples that contain attributable parts of answers.
Furthermore, while this dataset contains human-curated annotations of answers, it does not contains answer refusal examples. We augment Hagrid's training dataset with 25\% examples of answer refusals, sampling questions from Hagrid and random documents from Hagrid, Trust-Align, and ExpertsQA.

\item \textbf{ExpertsQA} \cite{malaviya2024expertqaexpertcuratedquestionsattributed} is a generative information retrieval dataset that includes human expert annotations and answers. Unlike traditional QA, it focuses on detailed answers, which are more challenging to attribute correctly due to their complexity and length.
Moreover, it is constructed across multiple domains using high-quality source documents, often from scientific or technical texts.

We preprocess the data by crafting examples that contain attributable parts of answers.
Then, similarly to the Hagrid dataset, we generate 25\% of examples where the model refuses to generate due to a lack of relevant information in the documents. To do so, we sample questions from ExpertsQA and random documents from Hagrid, \textit{Trust-Align}, and ExpertsQA.

\end{itemize}

We make sure that each example contains at least 5 retrieved documents. If an example contains less than 5, we add random documents sampled from the set of all documents (from the three datasets combined).
Note that we remove explicit citations (between brackets) from the text. In total, we obtain 20,312 training examples.

\subsection{Technical considerations}

\paragraph{\textbf{Marking strategy details}}

Regarding the marking tokens, our choice have been influenced by the Llama3.1 tokenizer. Our requirements were that: (1) tokens must be single entries in the tokenizer, and (2) the probability of encountering them in textual data during finetuning and evaluation is minimal. After examination, we settled on $10$ tokens that satisfy our requirements as document identifiers: [`` AA'', `` BB'', `` CC'', `` DD'', `` EE'', `` FF'', `` GG'', `` HH'', `` II", `` JJ''].
When marking a document, we randomly select a unique token identifier from this set.

To perform attribution, we aggregate the contributions across each sentence of the generated answer.
Furthermore, aligned with the trustworthiness objective, we add a fail-safe mechanism by replacing the answer with the refusal sentence described in \textit{Trust-Align} \cite{song2025measuringenhancingtrustworthinessllms} if no attribution is induced by our attribution method.

For fair comparison, in the main result, we use the top-5 documents retrieved for each query, provided by the ALCE benchmark.
Prompt-wise, we use a similar strategy as the \textit{Trust-Align} framework with a refusal clause, except that we do not ask the model to generate any citations.

\paragraph{\textbf{Identifier token logits learning}}
\label{sec:taininglabels}
The labels concerning the logit values to learn are set up empirically, based on training-validation on the Hagrid dataset \cite{kamalloo2023hagridhumanllmcollaborativedataset} as follows:
\begin{itemize}
    \item $\hat{l}_k = 4$ for document-identifier tokens in the ground truth attribution, i.e., $\{id_{k} \vert id_{k} \in a_i \} $. 
    \item $\hat{l}_k = 2$ for document-identifier tokens in the context but not in the ground truth attribution, i.e., $\{id_{k} \vert id_{k} \in C, id_{k}  \notin a_i\} $. 
    \item $\hat{l}_k = 0$ for document-identifier tokens of documents randomly sampled from $C$. 
    \item $\hat{l}_k = -2$ for document-identifier tokens of documents not in $C$ i.e., $\{id_{k} \vert id_{k}  \notin C\}$. 
\end{itemize}


Note that labels $\hat{l}_k = \{2, 4\}$ align with \textit{groundedness}, while labels $\hat{l}_k = \{0, -2\}$ align with \textit{context-relatedness}.

\paragraph{\textbf{Hyperparameters settings}}

We train using a learning rate of $2e-5$ with a cosine scheduler and an effective batch size of 16. We perform training for two epochs and use the resulting models to perform the evaluations.

The values of the $\alpha$ parameter (§ Eq. \ref{eq:mselogits}) in some of our losses were found empirically during preliminary experiments. We found that a value of $\alpha = 0.25$ gave the best results for every loss.

For each of the marking strategies in \textbf{LoDIT}, we investigated a static attribution threshold of $\phi_{\text{prop}} = 3$ ( § Eq. \ref{eq:aggrg}), which showed promises during early-stage experiments. We optimized the proportion threshold on the \textit{Hagrid} validation set, using the average of the F1GR and F1GC evaluation measures, as the set does not include gold-standard sub-strings or claims (present in ASQA, QAMPARI, and ELI5) to compute the F1AC measure.
Regarding $\lambda$ ( § Eq. \ref{eq:aggrg}), we investigate three values, $0.25$, $0.5$ and $0.75$ on validation and set $\lambda$ to $0.75$ for all results.


\section{Results and analysis}

\begin{table*}[]
\label{tab:tempresultstrusts}
\resizebox{\textwidth}{!}{
\begin{tabular}{lllcccccccccccc}
           &                           & \multicolumn{1}{l||}{}          & \multicolumn{4}{c|}{\textbf{ASQA}}                                                                       & \multicolumn{4}{c|}{\textbf{QAMPARI}}                                                                    & \multicolumn{4}{c}{\textbf{ELI5}}                                                         \\
             &                            & \multicolumn{1}{l||}{}          & F1AC                 & F1GR                 & F1GC                 & \multicolumn{1}{c|}{\textbf{TRUST}} & F1AC                 & F1GR                 & F1GC                 & \multicolumn{1}{c|}{\textbf{TRUST}} & F1AC                 & F1GR                 & F1GC                 & \textbf{TRUST}       \\ \hline 
\multicolumn{15}{c}{\textit{Baselines}}                                                                                                                                                                                                                                                                                                                                                                                \\ \hline
\multirow{4}{*}{Llama3 8B} & \multicolumn{1}{l}{\textit{PostCite}* \cite{gao2023enablinglargelanguagemodels}}         &            \multicolumn{1}{l||}{}                          & \multicolumn{1}{c}{32.98} & \multicolumn{1}{c}{53.31} & \multicolumn{1}{c}{28.01} & \multicolumn{1}{c|}{38.10}               & \multicolumn{1}{c}{6.10} & \multicolumn{1}{c}{34.52} & \multicolumn{1}{c}{8.42} & \multicolumn{1}{c|}{16.35}               & \multicolumn{1}{c}{20.80} & \multicolumn{1}{c}{45.88} & \multicolumn{1}{c}{8.06} & \multicolumn{1}{c}{24.91} \\ 
& \multicolumn{1}{l}{\textit{PostAttr}* \cite{gao2023enablinglargelanguagemodels}}     &    \multicolumn{1}{l||}{}                                           & \multicolumn{1}{c}{32.98} & \multicolumn{1}{c}{53.31} & \multicolumn{1}{c}{5.95} & \multicolumn{1}{c|}{30.75}               & \multicolumn{1}{c}{6.10} & \multicolumn{1}{c}{34.52} & \multicolumn{1}{c}{1.64} & \multicolumn{1}{c|}{14.09}               & \multicolumn{1}{c}{20.80} & \multicolumn{1}{c}{45.88} & \multicolumn{1}{c}{1.25} & \multicolumn{1}{c}{22.64} \\ 
& \multicolumn{1}{l}{Fine-tuned w/ \textit{FRONT}* \cite{huang2024learningfinegrainedgroundedcitations}}           &         \multicolumn{1}{l||}{}                                          & 62.25                & 41.62                & 66.14                & \multicolumn{1}{c|}{56.67}          & 13.53                & 22.78                & 20.42                & \multicolumn{1}{c|}{18.91}          & 18.99                & 17.85                & 44.69                & 27.18                \\
& \multicolumn{1}{l}{Fine-tuned w/ \textit{Trust-Align}* \cite{song2025measuringenhancingtrustworthinessllms}}   &   \multicolumn{1}{l||}{}                                           & 52.35                & 66.06                & 80.95                & \multicolumn{1}{c|}{66.45}          & 33.85                & 71.11                & 48.01                & \multicolumn{1}{c|}{\textbf{50.99}}          & 22.57                & 65.06                & 46.85                & 44.83                \\
\multirow{3}{*}{Llama3.1 8B} & $\boxplus$ \textit{MIRAGE} \cite{Qi_2024}    &           \multicolumn{1}{l||}{}                                            &       59.81          &       64.31       &     52.92           & \multicolumn{1}{c|}{59.01}          &        1.91       &         63.02      &         23.70      & \multicolumn{1}{c|}{29.54}          &        29.54     &        55.88       &       39.47        &     41.63       \\ 
& \multicolumn{1}{l}{Zero-Shot Prompting}    &         \multicolumn{1}{l||}{} 
&      \multicolumn{1}{c}{59.81} & \multicolumn{1}{c}{64.31} & \multicolumn{1}{c}{60.85} & \multicolumn{1}{c|}{61.66}               & \multicolumn{1}{c}{1.91} & \multicolumn{1}{c}{63.02} & \multicolumn{1}{c}{21.50} & \multicolumn{1}{c|}{28.81}               & \multicolumn{1}{c}{29.54} & \multicolumn{1}{c}{52.88} & \multicolumn{1}{c}{43.77} & \multicolumn{1}{c}{42.06} \\ 

& $\rhd$ Fine-tuned w/ \textit{Trust-Align} \cite{song2025measuringenhancingtrustworthinessllms}    &         \multicolumn{1}{l||}{} 
&       53.64          &       65.33       &     84.52            & \multicolumn{1}{c|}{67.83}          &        10.69        &        68.83       &         51.05        & \multicolumn{1}{c|}{43.53}          &       25.52        &       67.09         &        39.91        &       44.18      \\

\hline 
\multicolumn{15}{c}{\textit{Our markings and models}  }                                                                                                                                                                                                                                                                                                                                                                            \\ \hline
\multirow{6}{*}{Llama3.1 8B} & \multirow{3}{*}{Zero-Shot Prompting}    &    \multicolumn{1}{|l||}{$marking_{BA}$}                                        &         61.61             &         65.60             &          44.23            & \multicolumn{1}{c|}{57.15}               &          2.24            &          63.00            &          13.96            & \multicolumn{1}{c|}{26.40}               &          30.36            &            51.71          &           35.01           &           39.03           \\
&        &     \multicolumn{1}{|l||}{$marking_{BAS}$}                                    & \multicolumn{1}{c}{62.29} & \multicolumn{1}{c}{65.85} & \multicolumn{1}{c}{51.02} & \multicolumn{1}{c|}{59.72}               & \multicolumn{1}{c}{2.51} & \multicolumn{1}{c}{62.38} & \multicolumn{1}{c}{16.83} & \multicolumn{1}{c|}{27.24}               & \multicolumn{1}{c}{29.51} & \multicolumn{1}{c}{52.71} & \multicolumn{1}{c}{37.17} & \multicolumn{1}{c}{39.80} \\
&     &         \multicolumn{1}{|l||}{$marking_{AW}$}                                   & \multicolumn{1}{c}{60.36} & \multicolumn{1}{c}{62.40} & \multicolumn{1}{c}{36.65} & \multicolumn{1}{c|}{53.14}               & \multicolumn{1}{c}{2.06} & \multicolumn{1}{c}{62.22} & \multicolumn{1}{c}{12.54} & \multicolumn{1}{c|}{25.60}               & \multicolumn{1}{c}{29.65} & \multicolumn{1}{c}{53.09} & \multicolumn{1}{c}{27.00} & \multicolumn{1}{c}{36.58}    \\ \cline{2-15}
& \multirow{3}{*}{\textbf{LoDIT}} & \multicolumn{1}{|l||}{$marking_{BA}$}   & $61.52^{\blacktriangle\square}$                & $68.35^{\triangle\square}$                & $82.69^\blacksquare$                & \multicolumn{1}{c|}{$\mathbf{70.86^{\blacktriangle\blacksquare}}$} & $31.73^{\blacktriangle\blacksquare}$                 & $69.24^\blacksquare $               & $41.91^\blacksquare$                & \multicolumn{1}{c|}{$47.63^{\triangle \blacksquare}$}          & $27.98^\triangle$                 & $65.58^\blacksquare$                & $53.28^{\blacktriangle\blacksquare} $               & $\mathbf{48.94}^{\blacktriangle\blacksquare}$       \\
& &  \multicolumn{1}{|l||}{$marking_{BAS}$}   & $61.08^\blacktriangle$                & 63.76                & $71.56^\blacksquare$                & \multicolumn{1}{c|}{$65.47^\blacksquare$}          & $28.01^{\blacktriangle\blacksquare}$                & $68.24^\blacksquare$               & $30.68^\square$               & \multicolumn{1}{c|}{$42.27^\blacksquare$}          & 27.50                & $64.45^\blacksquare$               & $41.96^\triangle$                & $44.14^\square$                \\
& &  \multicolumn{1}{|l||}{$marking_{AW}$}  &54.76                & 63.44                & $81.11^\blacksquare$               & \multicolumn{1}{c|}{$66.44^\blacksquare$}          & $26.18^\blacktriangle$               & $68.02^\square $               & $34.55^\blacksquare$               & \multicolumn{1}{c|}{$42.91^\blacksquare$}          &  26.74                 & $60.15^\square$               & $49.23^{\blacktriangle\blacksquare}$                 & $46.23^{\triangle\blacksquare} $                \\ \hline
\end{tabular}
}
\caption{Comparison of \textbf{LoDIT} performance with state-of-the-art. Models with a star (*) have their performance measures reported from \cite{song2025measuringenhancingtrustworthinessllms}. $\blacktriangle$ and $\triangle$ symbols indicate significant improvements over the baseline model $\rhd$ Fine-tuned w/ \textit{Trust-Align} using a paired t-test with $\rho=0.01$ and $\rho=0.05$, respectively. $\blacksquare$ and $\square$ symbols indicate significant improvements over the baseline model $\boxplus$ \textit{MIRAGE} using a paired t-test with $\rho=0.01$ and $\rho=0.05$, respectively. Scores in bold are the best TRUST scores per dataset.\label{tab:mainresults}}
\end{table*}

\subsection{Main results}

Table \ref{tab:mainresults} presents the results of our evaluation on ASQA, QAMPARI, and ELI5 datasets. The table is split into baselines (top) and our marking and models (bottom).

Comparing \textbf{LoDIT} with the baselines, we can see on Table \ref{tab:mainresults} that \textbf{LoDIT} competes with the state-of-the-art, with $marking_{BA}$ consistently and significantly improving upon Llama 3.1 8B fine-tuned with \textit{Trust-Align} (which we will refer to as $\rhd$ \textit{Trust-Align}) and $\boxplus$ \textit{MIRAGE} in terms of TRUST score on all datasets, showing the overall strength of our proposal. As an example, we gain $4.76$ TRUST score over $\rhd$ \textit{Trust-Align} on the ELI5 dataset, corresponding to an improvement of $10.7\%$ ($48.94$ against $44.18$).
This strength is furthermore highlighted by the significant improvements achieved by \textbf{LoDIT} upon $\boxplus$ \textit{MIRAGE}, another approach leveraging models internals, in terms of F1GC on all datasets by up to $56.25\%$ ($82.69$ against $52.92$), $36.6\%$ ($41.91 $ against $30.68$), and $34.98\%$ ($53.28$ against $39.47$) on ASQA, QAMPARI, and ELI5, respectively.

Moreover, \textbf{LoDIT} outperforms Llama3 8B fine-tuned with \textit{Trust-Align}, on each evaluation metric for the ASQA and ELI5 datasets. For instance, we improve the TRUST score by $6.63\%$ ($70.86$ against $66.45$) on the ASQA dataset, and $9.96\%$ ($48.94$ against $44.83$) on the ELI5 dataset. Concerning the QAMPARI dataset, we hypothesize that the divergence in performance results from the difference in model used, as $\rhd$ \textit{Trust-Align} also shows weaker results on this dataset.

Focusing on the Zero-Shot Prompting  results, we see that using $marking_{BAS}$ outperforms other marking strategies on almost all the evaluation measures, with a clear win on the F1GC metric while the other two metrics remain similar. For instance, in the zero-shot scenario, $marking_{BAS}$ improves upon the second-best making strategy by up to $20\%$ ($16.83$ against $13.96$) in terms of F1GC on QAMPARI. 
However, analysing \textbf{LoDIT}, we see that $marking_{BA}$ outperforms other marking strategies, including all the Zero-Shot prompting scenarios.
This implies that a too refined marking might nonetheless impact performance when considering fine-tuning, and that a simpler one is more suited to the task, as highlighted by the state-of-the-art performance of \textbf{LoDIT} with $marking_{BA}$ marking.

\subsection{Ablation Study}

We conduct ablation studies to evaluate the effect of our debiaising logit strategy and  proposed aggregation function $\Psi$ (§ Eq.5). The scenario w/o, debiaising refers to a variant of \textbf{LoDIT} using the initial model-output logits. Concretely, the scenario w/o ,aggregation refers to a scenario where a variant naive aggregation is set up since raw token-based contribution values are needed. Thus, we consider the two naive following aggregation operators: (1) the \textit{max} strategy is defined as $a_i = \{id_k \vert \underset{k}{\arg\max}(ctr_{ij}^k) > \phi_{max}\}_{k=1}^{K}$ where  $\phi_{max}$ is a contribution threshold. The intuition behind \textit{max} approach is that documents should be attributed if they contribute highly to at least a single token \cite{Qi_2024} and it is somehow similar to max-pooling \cite{shen2018baselineneedslovesimple}; (2) the \textit{avg} strategy is defined as $a_i = \{id_k \vert avg[(ctr_{ij}^k)]_{j=1}^{n_i} > \phi_{avg}\}_{k=1}^{K}$ where $\phi_{avg}$ is a contribution threshold. The intuition behind \textit{avg} approach is that documents should be attributed if they contribute highly to all tokens, which is similar to mean-pooling \cite{reimers2019sentencebertsentenceembeddingsusing}. Results of the study are presented in Table \ref{tab:ablationstudy}.

We can see overall that the absence of logit-debiasing and k-pooling aggregations would result in a decline in performance, thereby assessing their joint relevant contribution to the performance of \textbf{LoDIT}. Particularly, we can see that the contribution of debiaising is relevant in scenarios involving each of the marking strategies and on each dataset, with an effect of its absence up to $-47.21\%$. Similarly, the ablation of our attribution strategy  using the  \textit{max} operator alternative deals with a range of negative effects between $0.69\%$ and $19.16\%$. However, we can notice that the \textit{avg} aggregation behaves in a dataset-dependent way. Indeed, ASQA and ELI5 may slightly benefit from using this operator for $marking_{BAS}$ and $marking_{AW}$, but not QAMPARI. Moreover, this ablation does not benefit our strongest marking strategy, namely $marking_{BA}$.

\begin{table}[]
\label{tab:ablationcontrib}
\resizebox{\columnwidth}{!}{%
\begin{tabular}{lll||c|c|c}
  &                                  &     & \textbf{ASQA}               & \textbf{QAMPARI}            & \textbf{ELI5}                  \\ \hline 
\multicolumn{2}{l}{$marking_{BA}$}   &     & 70.86                       & 47.60                       & 48.94                          \\
  & w/o debiaising                   &     & 55.13$^{\downarrow22.20\%}$ & 25.13$^{\downarrow47.21\%}$ & 30.01$^{\downarrow38.68\%}$    \\
  & \multirow{2}{*}{w/o aggregation} & + max & 67.76$^{\downarrow4.37\%}$  & 47.27$^{\downarrow0.69\%}$  & 46.65$^{\downarrow4.68\%}$     \\
  &                                  & + avg & 69.31$^{\downarrow2.19\%}$                   & 45.55$^{\downarrow4.31\%}$                   & 46.98$^{\downarrow4.00\%}$                      \\ \hline
\multicolumn{2}{l}{$marking_{BAS}$}  &     & 65.73                       & 39.39                       & 42.54                          \\
  & w/o debiaising                   &     & 54.64$^{\downarrow16.87\%}$ & 25.89$^{\downarrow34.27\%}$ & 31.68$^{\downarrow25.53\%}$    \\
  & \multirow{2}{*}{w/o aggregation} & + max & 60.17$^{\downarrow8.46\%}$  & 37.53$^{\downarrow4.72\%}$  & 37.97$^{\downarrow10.74\%}$    \\
  &                                  & + avg & 65.80$^{\uparrow0.11\%}$                   & 36.26$^{\downarrow7.95\%}$                   & 42.23$^{\downarrow0.73\%}$                      \\ \hline
\multicolumn{2}{l}{$marking_{AW}$}   &     & 68.05                       & 43.34                       & 47.13                          \\
  & w/o debiaising                   &     & 56.74$^{\downarrow16.62\%}$ & 26.17$^{\downarrow39.62\%}$ & 33.05$^{\downarrow29.87\%}$    \\
  & \multirow{2}{*}{w/o aggregation} & + max & 55.01$^{\downarrow19.16\%}$ & 42.85$^{\downarrow1.13\%}$  & 43.46$^{\downarrow7.79\%}$     \\
  &                                  & + avg & 66.07$^{\downarrow2.91\%}$                   & 40.94$^{\downarrow5.54\%}$                   &  47.98$^{\uparrow1.80\%}$  \\ \hline
\end{tabular}
}
\caption{Performance results (TRUST score) of the ablation study. $\downarrow$ and $\uparrow$ represent respectively the percentage decrease and percentage increase in TRUST score over \textbf{LoDIT} (w.r.t. to each marking strategy ).\label{tab:ablationstudy}}

\end{table}

\subsection{LoDIT Analysis}

\subsubsection{Logit debiaising}
\label{sec:Effectsoflogitdebiaising}

\begin{table}[]

\resizebox{\columnwidth}{!}{
\begin{tabular}{c||cc|cc|cc}
     & \multicolumn{2}{c|}{\textbf{ASQA}} & \multicolumn{2}{c|}{\textbf{QAMPARI}} & \multicolumn{2}{c}{\textbf{ELI5}} \\ 
     & \textbf{TRUST}      & Latency    & \textbf{TRUST}       & Latency       & \textbf{TRUST}      & Latency     \\ \hline 
$ablate-repeat$ &      50.36      &      771.14*       &     26.53        &        441.36*       &      29.99      &        734.20*     \\
$KL-proba$ &      62.01      &       \textbf{402.61}      &      39.44       &        \textbf{224.41}       &      43.18      &      381.74       \\ \hline
\textbf{LoDIT} &    \textbf{  70.86}      &      403.33       &      \textbf{ 47.63}      &       227.40        &     \textbf{ 48.94}      &  \textbf{375.94} \\ \hline        
\end{tabular}
}
\caption{Analysis of different logit debiasing strategies using $marking_{BA}$. Latency is reported in ms per query. * Note that parallel computing of the contributions may reduce the overhead without cancelling it completely. Bold scores indicate the highest value for the TRUST score and lowest value for the latency.\label{tab:debiasinglogits}}
\end{table}


To evaluate the potential of our logit-based debiasing stage, we explore two scenarios based on text ablation and probability debiasing, proposed in previous work \cite{Qi_2024, cohenwang2024contextciteattributingmodelgeneration}: (1) the $ablate-repeat$ scenario consists of ablating parts of the prompt (e.g., a document) and analyzing its effect on the model generation probabilities. The intuition being that the impact of a part of a prompt can be estimated by computing the model probability outputs difference between scenarios when the model is prompted with and without it.  
The authors use this difference in probabilities to infer which part of the prompt has to be attributed to the generated answer. In this case, the contribution calculation is redefined as the log difference \cite{cohenwang2024contextciteattributingmodelgeneration} between using and not using the context $C$:
\begin{equation}
\label{eq:probaIdC'dif}
ctr_{ij}^k=log \mathcal{M}(id_{k}\vert q, A_{S_{<i}},S_{i<j})-log \mathcal{M}(id_{k}\vert q, A_{S_{<i}},S_{i<j},C)
\end{equation}

(2) the $KL-proba$ scenario  consists of debiasing the token generation probabilities instead of their logits. We explore  the KL-divergence distillation loss \cite{hinton2015distillingknowledgeneuralnetwork}:

\begin{equation}
\mathcal{L}_{att}^{KL} = \sum_{ id_k\in a_i} KL(p_{\hat{l}_k} ||p(id_{k}\vert q, A_{S_{<i}},S_{i<j},C,\mathcal{M})) 
\end{equation}
where $p_{\hat{l}_k}$ is the generation probability associated with logit $\hat{l}_k$. Contributions are then computed using Eq. \ref{eq:ctr}.




Since latency overhead might be introduced by the scenarios, we report the average time in milliseconds used to perform a query on each evaluation dataset. Results in terms of performance and latency are presented in Table \ref{tab:debiasinglogits}, for both scenarios, as well as \textbf{LoDIT} using $marking_{BA}$ strategy. We can see that  the $KL-proba$ scenario outperforms the $ablate-repeat$ scenario in both metrics for the three datasets. This result indicates that the $ablate-repeat$ can be easily outperformed, in terms of quality of the attributions but also in terms of latency, as indicated by an almost doubling of the latency on all the datasets, as well as drops in terms of TRUST score of $28.93\%$ ($70.86$ vs. $50.36$), $44.29\%$ ($47.63$ vs. $26.53$) and $38.72\%$ ($48.94$ against $29.99$) on ASQA, QAMPARI and ELI5 when compared with \textbf{LoDIT}.
However, when comparing the logit-based strategy used in \textbf{LoDIT} and $KL-proba$, there are no strong differences in terms of latency, but TRUST performance scores are clearly different, with our logit-based strategy outperforming in the three datasets, highlighted by improvements of $14.27\%$ ($70.86$ vs.  $62.01$), $20.76\%$ ($47.63$ vs. $39.44$), and $13.33\%$ ($48.94$ against $43.18$) on ASQA, QAMPARI, and ELI5, respectively.

\subsubsection{Robustness analysis of \textbf{LoDIT}}

Since \textbf{LoDIT} is based on marked documents in the context, we analyse how impactful the number and order of these documents in the context are on its performance.

\begin{figure}
\label{fig:fignbdocs}
\resizebox{0.9\columnwidth}{!}{
\begin{tikzpicture}
\begin{axis}[
    ybar,
    bar width=10pt,
    width=12cm, height=7cm,       
    xtick={0,1,...,10},
    legend style={at={(0.78,1.17)},anchor=north,legend columns=-1}, 
    symbolic x coords={2, 4, 5, 6, 8, 10}, 
    xtick=data,                   
    ylabel={TRUST score},           
    xlabel={Length of context in terms of number of documents (K)},             
]

\addplot[style={fill=lightgray},
    thick] 
    coordinates {(2, 62.63) (4, 68.68) (5, 70.86) (6, 70.41) (8, 67.19) (10, 64.91)};

\addplot[style={fill=purple},
    thick] 
    coordinates {(2, 40.63) (4, 46.08) (5, 47.63) (6, 44.96) (8, 44.58) (10, 45.00)};

\addplot[    style={fill=teal},
    thick] 
    coordinates {(2, 46.83) (4, 46.27) (5, 48.94) (6, 47.35) (8, 42.85) (10, 45.54)};

\legend {ASQA, QAMPARI, ELI5}

\end{axis}

\draw[dashed, darkgray, thick] (4,4.95) -- (10.4,4.95);
\draw[dashed, darkgray, thick] (0.5,3.75) -- (10.4,3.75);

\draw[dashed, purple, thick] (0,1.5) -- (4.5,1.5);
\draw[dashed, purple, thick] (0,0.44) -- (1,0.44);

\draw[dashed, teal, thick] (4.6,1.68) -- (10.4,1.68);
\draw[dashed, teal, thick] (8.5,0.78) -- (10.4,0.78);

\end{tikzpicture}}
\vspace{-0.3cm}
\caption{Robustness of \textbf{LoDIT} to the length of context.\label{fig:RobustnessNumDocs}}
\vspace{-0.5 cm}
\end{figure}
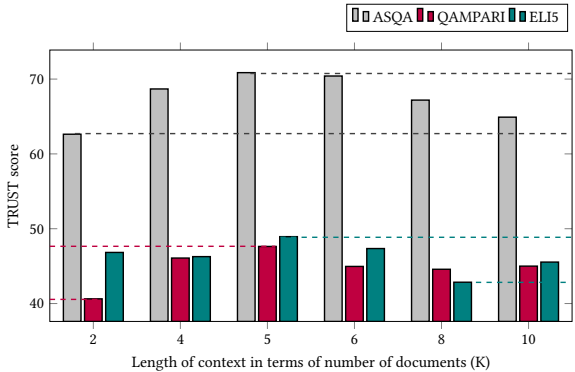

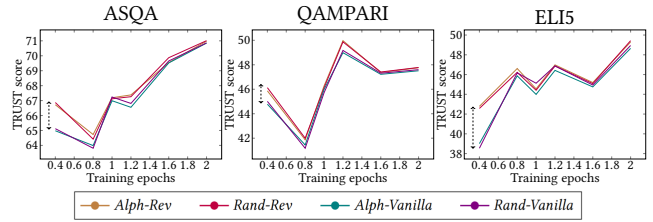
\begin{figure}[h]
\centering
\resizebox{\columnwidth}{!}{

\begin{tikzpicture}
\begin{axis}[
    title={ASQA},
    xlabel={Training epochs},
    ylabel={TRUST score},
    width=12cm,
    height=9cm,
    mark=*,
    legend style={draw=none},
    legend entries={},
    label style={font=\fontsize{22}{22}\selectfont},          
    tick label style={font=\fontsize{22}{22}\selectfont},
    title style={font=\fontsize{35}{35}\selectfont}
]

\addplot[
    color=brown,
    thick
] table[row sep=\\] {
    X Y \\
    0.4 66.72 \\
    0.8 64.73 \\
    1.0 67.19 \\
    1.2 67.38 \\
    1.6 69.54 \\
    2 70.98 \\
};
\addlegendentry{\textit{Alph-Rev}}

\addplot[
    color=purple,
    thick
] table[row sep=\\] {
    X Y \\
    0.4 66.87 \\
    0.8 64.42 \\
    1.0 67.12 \\
    1.2 67.26 \\
    1.6 69.87 \\
    2 71.01 \\
};
\addlegendentry{\textit{Rand-Rev}}

\addplot[
    color=teal,
    thick
] table[row sep=\\] {
    X Y \\
    0.4 64.98 \\
    0.8 63.99 \\
    1.0 66.99 \\
    1.2 66.55 \\
    1.6 69.52 \\
    2 70.85 \\
};
\addlegendentry{\textit{Alph-Vanilla}}

\addplot[
    color=violet,
    thick
] table[row sep=\\] {
    X Y \\
    0.4 65.12 \\
    0.8 63.81 \\
    1.0 67.24 \\
    1.2 66.81 \\
    1.6 69.65 \\
    2 70.86 \\
};
\addlegendentry{\textit{Rand-Vanilla}}
\legend{};
\end{axis}
\draw[dashed, black, thick, <->] (0.50,1.7) -- (0.50,3.3);
\end{tikzpicture}

\begin{tikzpicture}
\begin{axis}[
    title={QAMPARI},
    xlabel={Training epochs},
    ylabel={TRUST score},
    width=12cm,
    height=9cm,
    mark=*, 
    legend pos=south east,
    label style={font=\fontsize{22}{22}\selectfont},          
    tick label style={font=\fontsize{22}{22}\selectfont},
    title style={font=\fontsize{35}{35}\selectfont}
]
\addplot[
    color=brown,
    thick
] table[row sep=\\] {
    X Y \\
    0.4 45.87 \\
    0.8 41.87 \\
    1.0 46.27 \\
    1.2 49.97 \\
    1.6 47.37 \\
    2 47.76 \\
};
\addlegendentry{\textit{Alph-Rev}}

\addplot[
    color=purple,
    thick
] table[row sep=\\] {
    X Y \\
    0.4 46.13 \\
    0.8 42.00 \\
    1.0 46.12 \\
    1.2 49.87 \\
    1.6 47.41 \\
    2 47.78 \\
};
\addlegendentry{\textit{Rand-Rev}}

\addplot[
    color=teal,
    thick
] table[row sep=\\] {
    X Y \\
    0.4 44.78 \\
    0.8 41.45 \\
    1.0 45.98 \\
    1.2 48.99 \\
    1.6 47.22 \\
    2 47.51 \\
};
\addlegendentry{\textit{Alph-Vanilla}}

\addplot[
    color=violet,
    thick
] table[row sep=\\] {
    X Y \\
    0.4 45.01 \\
    0.8 41.18 \\
    1.0 45.72 \\
    1.2 49.17 \\
    1.6 47.31 \\
    2 47.60 \\
};
\addlegendentry{\textit{Rand-Vanilla}}
\legend{};
\end{axis}

\draw[dashed, black, thick, <->] (0.50,3.2) -- (0.50,4.3);
\end{tikzpicture}
\begin{tikzpicture}
\begin{axis}[
    title={ELI5},
    xlabel={Training epochs},
    ylabel={TRUST score},
    width=12cm,
    height=9cm,
    mark=*, 
    legend pos=south east,
    label style={font=\fontsize{22}{22}\selectfont},          
    tick label style={font=\fontsize{22}{22}\selectfont},
    title style={font=\fontsize{35}{35}\selectfont}
]
\addplot[
    color=brown,
    thick
] table[row sep=\\] {
    X Y \\
    0.4 42.78 \\
    0.8 46.61 \\
    1.0 44.55 \\
    1.2 46.98 \\
    1.6 45.19 \\
    2 49.30 \\
};
\addlegendentry{\textit{Alph-Rev}}

\addplot[
    color=purple,
    thick
] table[row sep=\\] {
    X Y \\
    0.4 42.57 \\
    0.8 46.22 \\
    1.0 44.41 \\
    1.2 46.87 \\
    1.6 45.08 \\
    2 49.43 \\
};
\addlegendentry{\textit{Rand-Rev}}

\addplot[
    color=teal,
    thick
] table[row sep=\\] {
    X Y \\
    0.4 39.01 \\
    0.8 45.87 \\
    1.0 43.99 \\
    1.2 46.42 \\
    1.6 44.75 \\
    2 48.65 \\
};
\addlegendentry{\textit{Alph-Vanilla}}

\addplot[
    color=violet,
    thick
] table[row sep=\\] {
    X Y \\
    0.4 38.54 \\
    0.8 46.17 \\
    1.0 45.12 \\
    1.2 46.87 \\
    1.6 44.94 \\
    2 48.94 \\
};
\addlegendentry{\textit{Rand-Vanilla}}
\legend{};
\end{axis}
\draw[dashed, black, thick, <->] (0.50,0.6) -- (0.50,3.0);
\end{tikzpicture}
}
\includegraphics[width=0.8\columnwidth]{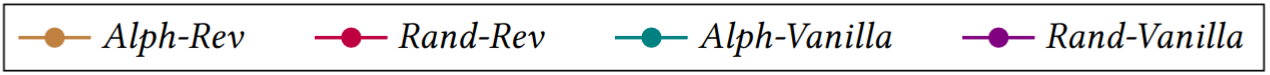}

\caption{Robustness of \textbf{LoDIT}  to document ordering and identifier token selection. \textit{Rand-Vanilla} (purple) corresponds to the default configuration of our $marking_{BA}$ strategy.\label{fig:docorder}}

\end{figure}

\paragraph{\textbf{Context length analysis}}
Figure \ref{fig:RobustnessNumDocs} shows our robustness analysis regarding the number of documents used in the context. We perform evaluations using the $marking_{BA}$ strategy with the top $K$ documents with $K$ in the range $[2,10]$ by step of 2. As a reminder, \textbf{LoDIT} is trained using $5$ top-retrieved documents in the context.
Unsurprisingly, we notice that for each dataset, the best performing configuration is using $5$ documents, with performance dropping as we increase or decrease the number of documents. These results show that the best performance is directly linked to the configurations seen during training, and encourage a more diverse training procedure to increase robustness. 
Nonetheless, \textbf{LoDIT} remains robust to slight changes in the number of documents as highlighted by drops of TRUST score of only $0.63\%$ and $3.10\%$ on ASQA, $5.26\%$ and $3.21\%$ on QAMPARI, $3.28\%$ and $5.29\%$ on ELI5 when comparing using 5 documents to using 6 and 4, respectively.
\paragraph{\textbf{Document order analysis}}
Figure \ref{fig:docorder} shows the performance results of different configurations regarding the order of documents in the context as well as the choice of identifier tokens during inference. More precisely, it highlights the TRUST performance score (on the test data) at different steps of the training. 
We define \textit{Rand}, as the configuration where we randomly assign identifier tokens to documents during the marking step, and \textit{Alph}, as the scenario where  identifier tokens are alphabetically ordered during marking (first $`` AA"$, then $`` BB"$, etc.). We define \textit{Vanilla} as the traditional document ranking of documents  in the context based on their relevance to the input query, and \textit{Rev}, which is the reverse of the relevance-based ranking. We analyse each combination of token marking selection and document ordering configuration to evaluate their joint effect on performance (e.g., \textit{Rand-Vanilla}, which is the ``default'' configuration of \textbf{LoDIT}).
By  comparing \textit{Alph-Vanilla} with \textit{Rand-Vanilla} and \textit{Alph-Rev} with \textit{Rand-Rev}, we can conclude from Figure \ref{fig:docorder} that \textbf{LoDIT} is robust to the choice of identifier tokens as early as 40\% of the first epochs for each dataset. That is explained by the fact that \textbf{LoDIT} is trained using random choosing of identifier tokens and the alphabetical ordering simply being a special case of random ordering.
Furthermore, by analyzing the order  of documents (i.e., by comparing \textit{Alph-Vanilla} with \textit{Alph-Rev} and \textit{Rand-Vanilla} with \textit{Rand-Rev}), we first see that ordering the documents in reverse of relevance-based ranking slightly improves the performance during the early stages of training. Second, we see that this improvement reduces as training progresses, with performance in all the evaluation configurations converging during the late stage of training. This observation highlights that \textbf{LoDIT} learns to more accurately consider documents regardless of their position in the context, which advocates for its robustness.

\subsubsection{Proportion threshold analysis}
The aggregation operator uses the hyperparameter, $\lambda$ (§ Eq.5), representing the proportion of statements to be considered to perform attribution. While we set this hyperparameter to $\lambda = 0.75$ for all experiments after an optimization on the Trust-Align test set, we analyze here different values of this parameter and their impact of the TRUST score.

Table \ref{tab:propanalysis} shows the TRUST score of \textbf{LoDIT} with the $marking_{BA}$ strategy on the evaluation datasets for three values of $\lambda$.
We notice that for all datasets, the TRUST score is only slightly impacted with respect to $\lambda$. We can see that the TRUST score loss percentage between $\lambda = 0.75$ and $\lambda=0.25$ is $2.35\%$ ($70.86$ against $69.19$), $3.40\%$ ($47.63$ against $46.01$) and $1.63\%$ ($48.94$ against $48.14$) for ASQA, QAMPARI and ELI5 respectively.
We argue that these results show the confidence of \textbf{LoDIT} in its debiaising approach, giving ``high'' logit values to the same identifier tokens for each generated token in the statement. In other words, if the logits' values of the identifier tokens are similar at each step of the generation, the effects of considering whether $25\%$, $50\%$ or $75\%$ of these values are higher than a threshold is negated.

\begin{table}[h]
\centering

\begin{tabular}{c||c|c|c}
\textbf{Proportion} & \multirow{2}{*}{\textbf{ASQA}} & \multirow{2}{*}{\textbf{QAMPARI}} & \multirow{2}{*}{\textbf{ELI5}} \\
\textbf{ Threshold ($\lambda$)} &  & & \\
\hline 
0.25 & 69.19 & 46.01 & 48.14 \\
0.50 & 69.85 & 46.78 & 49.16 \\
0.75 & 70.86 & 47.63 & 48.94 \\
\hline
\end{tabular}
\caption{Impact of the proportion threshold on the Trust score when using the $marking_{BA}$ strategy.}\label{tab:propanalysis}
\end{table}

\vspace{-0.6cm}
\vspace{-0,3 cm}

\section{Conclusion}
In this work, we propose leveraging identifiers of documents in a retrieved context  to faithfully attribute LLM-generated answers using a RAG framework. To bridge the gap between model generation and attribution, we propose  a two-level attribution method, called \textbf{LoDIT}, where the document identifier logit is used to estimate token-level contributions, which are then aggregated to compute the statement-level attribution. Extensive experiments using the recent Trust-Align evaluation framework validate the effectiveness, efficiency, and robustness of \textbf{LoDIT}.\\
An interesting avenue of research involves extending \textbf{LoDIT} toward an end-to-end learnable token-level and statement-level attribution. Additionally, investigating how our logit-based approach could be extended to answer token selection and mitigate other types of bias would allow us to enhance the fairness of model outputs, beyond faithfulness and trustworthiness.

\end{sloppypar}
\bibliographystyle{plain}
\bibliography{paper}
\end{document}